\documentclass[10pt, conference, compsocconf]{IEEEtran}

\usepackage{booktabs} 
\usepackage{subfig}
\usepackage{multirow}
\usepackage{graphics}
\usepackage{graphicx}
\usepackage{array}
\usepackage{xspace}
\usepackage{url}
\usepackage{color}

\usepackage[cmex10]{amsmath}
\usepackage{amssymb}
\usepackage{array}

\newcommand{\tuple}[1]{\langle#1\rangle}
\newcommand{\mi}[1]{\mathit{#1}}
\newcommand{\cR}{\ensuremath{\mathcal{R}}\xspace}
\newcommand{\cE}{\ensuremath{\mathcal{E}}\xspace}
\newcommand{\seq}{\ensuremath{\mathcal{S}}\xspace}

\newcommand{\cL}{\ensuremath{\mathcal{L}}\xspace}
\newcommand{\cX}{\ensuremath{\mathcal{X}}\xspace}
\newcommand{\kg}{\ensuremath{\mathcal{K}}\xspace}

\newcommand{\EKLSkip}{EKL$_{\textit{Skip}}$\xspace}
\newcommand{\EKLConcat}{EKL$_{\textit{Concat}}$\xspace}
\newcommand{\EKLRNN}{EKL$_{\textit{RNN}}$\xspace}

\begin{document}

\title{
On Event-Driven Knowledge Graph Completion in Digital Factories}

%

%






\author{
\IEEEauthorblockN{Martin Ringsquandl}
	\IEEEauthorblockA{
		Ludwig-Maximilians Universit{\"a}t \\
		Munich, Germany\\
		\small{martin.ringsquandl@gmail.com}}
	\and
	\IEEEauthorblockN{Evgeny Kharlamov}
	\IEEEauthorblockA{
		Oxford University \\
		Oxford, United Kingdom\\
		\small{evgeny.kharlamov@cs.ox.ac.uk}}
	\and
	\IEEEauthorblockN{Daria Stepanova}
	\IEEEauthorblockA{
		Max-Planck-Institut f\"ur Informatik \\
		Saarbr\"ucken, Germany \\
		\small{dstepano@mpi-inf.mpg.de}}
	\and
	\IEEEauthorblockN{Steffen Lamparter}
	\IEEEauthorblockA{
		Siemens AG CT \\
		Munich, Germany \\
		\small{steffen.lamparter@siemens.com}}
	\and
	\IEEEauthorblockN{Raffaello Lepratti}
	\IEEEauthorblockA{
		Siemens PLM Software  \\
		Genoa, Italy \\
		\small{raffaello.lepratti@siemens.com}}
	\and
	\IEEEauthorblockN{Ian Horrocks}
	\IEEEauthorblockA{
		Oxford University \\
		Oxford, United Kingdom\\
		\small{ian.horrocks@cs.ox.ac.uk}}
	\and
\IEEEauthorblockN{Peer Kr\"oger}
	\IEEEauthorblockA{
		Ludwig-Maximilians Universit{\"a}t \\
		Munich, Germany \\
		\small{kroeger@dbs.ifi.lmu.de}}
}


\maketitle

\begin{abstract}
Smart factories are equipped with machines that can sense their manufacturing environments, interact with each other, and control production processes. Smooth operation of such factories requires that the machines and engineering personnel that conduct their monitoring and diagnostics share a detailed common industrial knowledge about the factory, e.g., in the form of knowledge graphs. Creation and maintenance of such knowledge is expensive and requires automation. In this work we show how machine learning that is specifically tailored towards industrial applications can help in knowledge graph completion. In particular, we show how knowledge completion can benefit from event logs that are common in smart factories. We evaluate this on the knowledge graph from a real world-inspired smart factory with encouraging results.
\end{abstract}


\IEEEpeerreviewmaketitle


\section{Introduction}
\label{sec:introduction}

\subsection{Motivation}
Digitalisation and automation are among the biggest trends in manufacturing~\cite{Rosen2015}. 
Modern automated, or \emph{smart}, factories are equipped with production and assembling machines that are not only capable of \emph{sensing} their environments, 
e.g. reading RFID tags of products, 
but also of \emph{interacting} with each other, e.g. raising a material shortage warning, and even performing \emph{controlling} actions, e.g. turning on a cooling fan. 
Thus, it is common to distinguish in a smart factory its \emph{physical part}, that is, machines and production lines, and its \emph{digital representation}, referred to as the \textit{digital twin}~\cite{Datta2016}.

\begin{figure}
	\centering
	\includegraphics[width=.4\textwidth]{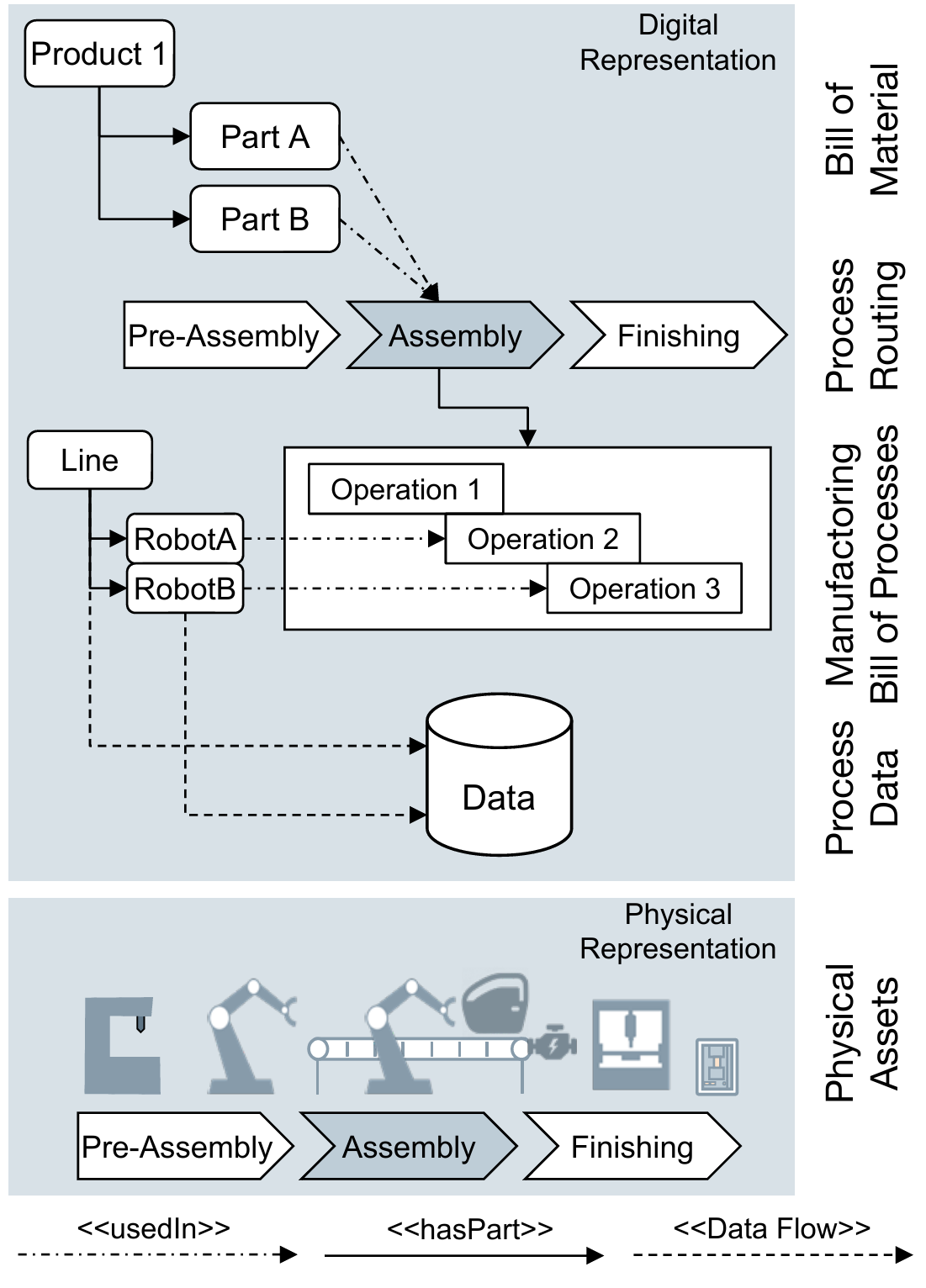}
	\caption{Schematic split in physical and digital representation of a factory}
	\label{fig:digital_twin}
\end{figure}

The digital twin acts as an interface to the physical system, 
offering services such as automated monitoring, optimisation, and ultimately self-organisation without the need to interact with its actual physical representation~\cite{Hirmer2017}.
%
%
In Figure~\ref{fig:digital_twin} we schematically visualise the separation between the physical part (in the bottom) and the digital twin (on top). 
The physical part consists of several machines for pre-assembly, assembly, and finishing of manufacturing.
The digital twin consists of a specification of 
\emph{Product1} that says that the product has two components \emph{PartA} and \emph{PartB} and can be assembled with three operations,
where the last two are conducted by robots \emph{RobotA} and \emph{RobotB} that in turn are located in a manufacturing line.


Development and maintenance of a digital twin poses significant challenges. 
In particular, one has to ensure that the relevant \emph{industrial knowledge} about the plant is well captured and maintained.
This knowledge representation is in the heart of the digital twin, upon which 
applications are built that rely and refer to it as backbone for communication.
The knowledge should encompass both \emph{master} and \emph{operational data} (which are partially depicted in Figure~\ref{fig:digital_twin}).
The former includes the catalog of plant's equipment together with its technical documentation and the topology of its location in the manufacturing shop floor, personnel, warehouse data, and production blueprints.
The latter includes log files of messages generated by individual pieces of equipment during manufacturing, flow of raw materials and products, and purchases.

Such industrial knowledge naturally satisfies the Big Data dimensions. Indeed, first,
it is large in \emph{volume}, 
e.g., at a mid size plant this knowledge may contain information about up to hundreds of machines, processes and materials, and hundreds of thousands of events.
Second, the data \emph{velocity} is high, e.g.,
a daily volume of transaction data generated only by shop-floor equipment can be up to hundreds of thousands of messages,
master data is also dynamic in this regard: shop-floor devices may be added, moved, or removed due to maintenance, system configurations may change according to the respective production processes and products when, say, a new product variant requires an additional welding operation.
Finally, 
the \emph{variety} across various data sources is high,
e.g., the transaction data is structured according to numerous relational schemata,
technical documentation comes in flat files, 
and equipment capabilities are encoded in various proprietary formats.

\emph{Knowledge Graphs} (KGs) are considered as a prominent approach 
to represent and share industrial knowledge~\cite{Kharlamov2014,Ringsquandl2015,OptiqueStatoil,OptiqueSiemens,DBLP:journals/internet/HorrocksGKW16}
since they offer a flexible mechanism for structuring both master and transaction data as an interconnected network of entities.
Knowledge graphs are typically either 
available or can be exported
as W3C standardised RDF datasets%
\footnote{\url{http://www.w3.org/RDF/}}
that consist of \emph{triples},
e.g., of the form $\mi{\tuple{entity,predicate,entity}}$.
This format is well suited for both knowledge representation and exchange across applications over the network.

%

A typical KG in a digital factory consists of several logical parts
that capture the main components of a digital twin in a smart factory~\cite{Kharlamov2016a} and can be found in Figure~\ref{fig:digital_twin}:
\emph{Bill of Material} (i.e. a partonomy of products and materials), \emph{Process Routings} (i.e. sequences of processes), the \emph{Manufacturing Bill of Process} (i.e. assignment of machines to processes and more detailed operations), and \emph{Process Data} (i.e. data collected from the machines during production). When equipped with rich semantic descriptions, these entities and their relationships are  what constitutes the digital representation of the factory.


\subsection{Challenges with Knowledge Graphs}
Creating and maintaining a KG of good quality is a challenging task
and an important bottleneck for the adoption of digital twins in industry~\cite{Ringsquandl.2017}:
due to the Big Data dimensions of industrial knowledge, 
the corresponding KG cannot be manually created and curated.
Thus, semi-automated techniques are needed to create and expand industrial KGs and a number of machine learning techniques have been proposed to address this challenge~(see, e.g., \cite{Nickel2015} for overview).
The main idea of these approaches is to convert entities and relations in a KG into a low-dimensional vector space and 
use it to infer missing information in the KG.

These approaches work better when the vector space is enhanced with some extra background knowledge,
e.g., by also embedding textual documents attached to entities. 
We refer the reader to Section~\ref{sec:approach} for more details on existing machine leaning approaches for KG expansion.

\subsection{Contributions}
In this work we show the effectiveness of  
machine learning for knowledge graph completion where
the learning method is based on the vector space representation and accounts for background knowledge. 
In particular we develop an industrial scenario of a smart factory, a knowledge graph describing this factory, and how completion of this graph with the help of machine learning  can be enhanced with a special type of background knowledge: log files of event messages generated by shop-floor manufacturing equipment.

The results of our evaluation are encouraging, where we apply the machine learning models to an industrial KG containing roughly 6,000 facts and create several use case scenarios of missing information. We show that our approach yields a boost in the quality of predicting missing links between digital twin entities and for certain relations  we are able to restore missing information even in scenarios with highly incomplete relations.

\section{Use Case: Incomplete Information in a Dynamic Manufacturing System}
\label{sec:usecase}

Our use case scenario is focused on synchronizing \textit{Digital Twins} with their physical counterparts, more precisely we focus on a digital representation of automated factories that exhibit missing information. At the heart of such a digital factory representation is a data model describing the automation equipment, i.e. controllers and actuators, and other entities typically found in manufacturing environments, such as processes, products, and events. 

The rest of the section is organised as follows.
In Section~\ref{section:Factory Description} we describe the factory we study in our use case. 
In Section~\ref{section:KGs} we explain how we turn factory data in a KG.
In Section~\ref{section:events} we describes event data we collected and why this data is relevant for our use case.
Then, we present three scenarios of missing information in industrial knowledge graphs that we investigate in this work.
All scenarios correspond to real-world situations in a factory that can be observed in smart factory environments.
The scenarios are: 
change of factory equipment (Section~\ref{ssec:equip}), introduction of 
new processes in manufacturing (Section~\ref{ssec:process}),
and update of equipments software that results in new events being emitted (Section~\ref{ssec:events}).

\subsection{Factory Description}
\label{section:Factory Description}

The factory we studied is a simplified simulation of a real-world smart factory, and it consists of four similarly structured production lines, each of which produces a particular variant of a common product family using a set of connected production equipment. The factory has 180 devices, 4 different products that consist of a total of 59 unique material parts, and 55 different manufacturing processes. Each device can perform some skills including drilling, welding, and assembling
and operates by inputting and outputting some of the materials that are part of four different product variants. In total the devices emitted 728 unique event entities during the collection time period for this case study, more details on the event data are given in the subsequent section.

\subsection{Manufacturing Data as a Knowledge Graph}
\label{section:KGs}

Typically, manufacturing data is scattered throughout diverse data sources and formats (relational databases, spreadsheets, XML files). Since we rely on RDF-based knowledge graphs,
we exploit an ontology driven ETL process, known as \emph{Ontology-Based Data Access} (\emph{OBDA}) in order to translate these heterogeneous data sources into RDF.

OBDA follows the classical data integration para\-digm 
	that requires the creation of a common `global' schema
	that consolidates `local' schemata of the integrated data sources,
	and mappings 
	that define how the local and global schemata are related~\cite{DBLP:books/daglib/0029346}.
	In OBDA the global schema is an \emph{ontology}:
	a formal conceptualisation of the domain of interest 
	that consists of a \emph{vocabulary}, i.e., names
	of classes, attributes and binary relations, 
	such as \textit{connectedTo}, \emph{hasPart},
	and \emph{axioms} over the terms from the vocabulary
	that, e.g., assign attributes of classes,
	define relationships between classes, 
	compose classes, class hierarchies, etc.
The ontology we developed
	encodes 
	generic specifications of manufacturing equipment, 
	characteristics of sensors, 
	materials,
	processes,
	descriptions of diagnostic tasks, etc.
OBDA mappings relate each ontological term
	 to a set of queries over the underlying data.
OBDA mappings can be used in the same way as ETL rules for data transformation.

%

An overview of the KG that we obtained with the help of OBDA is shown in Table \ref{tb:datasets}, where $|\cE_c|$ is the number of entities in the given class, $avg(In)$ represents the average number of incoming, and $avg(Out)$ the average number of outgoing links for each of the entity classes. Note that equipment entities are most densely connected in the KG, while events, although the largest class of entities in the KG, only have few outgoing links.
	\begin{table}
		\begin{center}
			\caption{Main entities in the digital twin knowledge graph}
			\label{tb:datasets}
			\begin{tabular}{| p{2cm} | >{\centering\arraybackslash}m{1cm} | >{\centering\arraybackslash}m{2cm} |
			> {\centering\arraybackslash}m{2cm} | }
				\hline
				\textbf{Entity Class} & $|\cE_c|$ & $avg(In)$ & $avg(Out)$ \\ \hline
				Equipment & $180$ & $13.13$ & $5.6$ \\ \hline
				Process & $55$ & $4.89$ & $7.0$ \\ \hline
				Material & $59$ & $5.9$ & $8.9$ \\ \hline
				Event & $728$ & $0$ & $2.17$ \\ \hline
			\end{tabular}
		\end{center}
	\end{table}
In summary, the KG consists of 3,125 entities that are connected through 6,361 facts (triples) in 11 unique named relations (predicates).


\subsection{Event Data}
\label{section:events}
Factories, such as the one we simulated, are equipped with automation systems that continuously generate operational data, especially events, such as alarm codes or operator information messages. For a particular point in time, events from different locations in the factory are collected and later aggregated in log files. Observing these events enables the digital twin to trace some of the activity that is carried out by the physical equipment.

Due to the scattered layout of machines across factories, two consecutively emitted events are not necessarily correlated to each other, since their physical sources (i.e. production machines) may be completely independent. Nevertheless, as we will show in the following sections, co-occurrence patterns in event logs can be used to infer missing information contained in the factories KG.

For our use case study, we collected about 60,000 occurrences of events from the simulated factory.

\subsection{Scenario: Changing Factory Equipment}
\label{ssec:equip}

The first scenario of completing missing information in the KG that we consider is related to equipment entities. More precisely, we study the effectiveness of KG completion by artificially removing certain links between equipment entities. Such missing information can naturally occur when additional devices are deployed at the factory, or existing devices need to be replaced due to maintenance. Having background knowledge in form of event sequences, this scenario studies how well such equipment connections can be automatically re-established through inference.

\begin{figure}[t!]
	\includegraphics[trim=0cm 0cm 0cm 0cm, clip, width=0.49\textwidth]{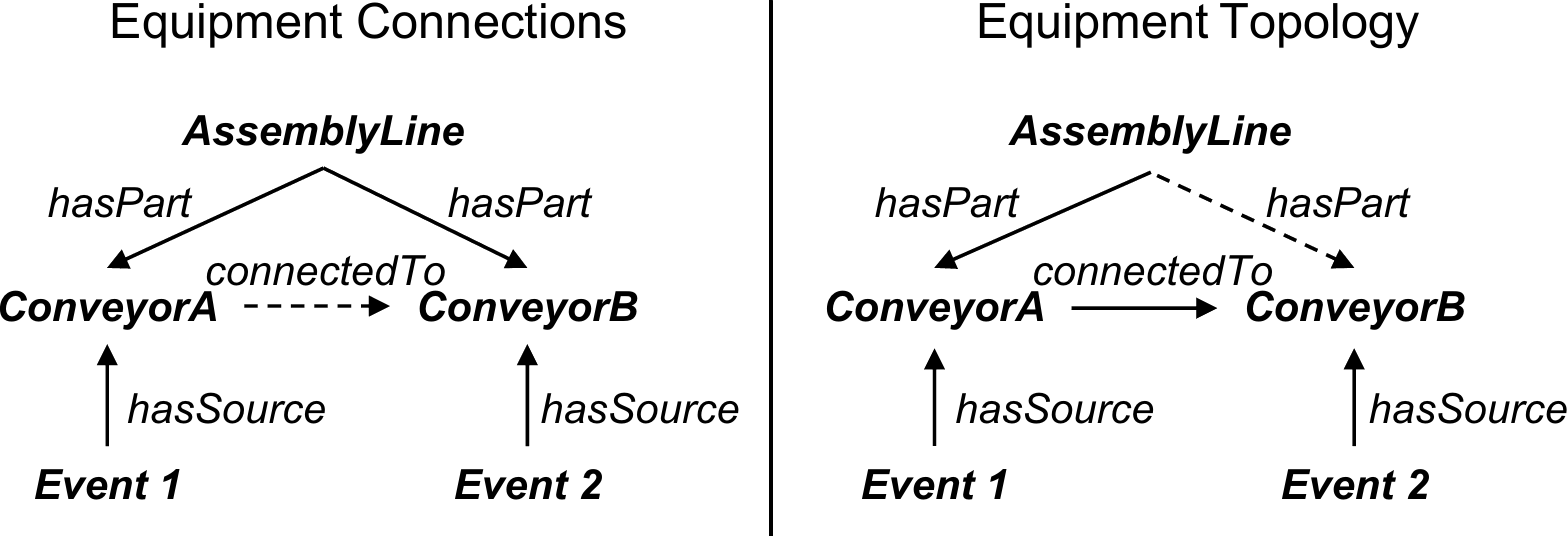}
	\caption{Equipment scenario, left: missing links that state facts about physical device connections. Right: Missing partonomy links of equipment}
	\label{fig:equip}
\end{figure}

For our manufacturing KG, this scenario mainly affects two relations \textit{hasPart} and \textit{connectedTo}, as shown in Figure \ref{fig:equip}. Both sides of the figure show the same KG consisting of an assembly line that has two conveyors as parts. Also two exemplary event entities are related to their respective source locations. On the left-hand side of the figure the \textit{connectedTo} relation is artificially removed, as indicated by the dashed arrow. On the right-hand side one of the \textit{hasPart} relations is removed.

The KG completion task is to obtain a correct recommendation for both types of missing links, such that the missing information is restored via inference.

\subsection{Scenario: Introduction of New Processes}
\label{ssec:process}

\begin{figure}[t!]
	\includegraphics[trim=0cm 0cm 0cm 0cm, clip, width=0.49\textwidth]{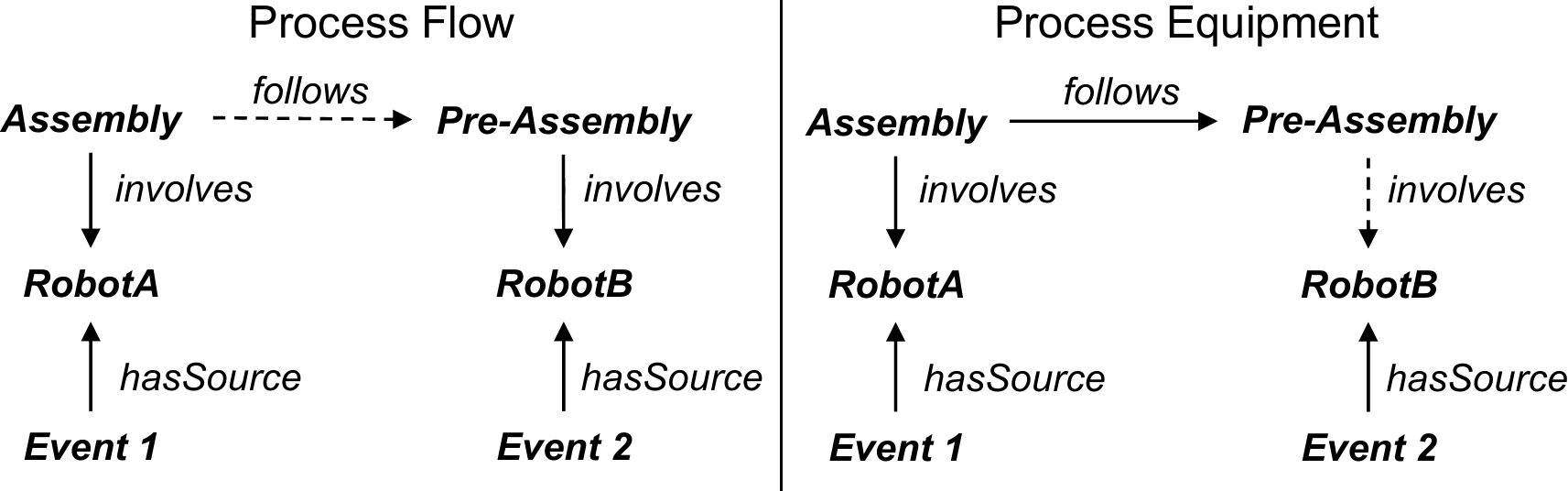}
	\caption{Process scenario, left: missing links that state facts the production process flow. Right: Missing links to involved equipment in the process}
	\label{fig:process}
\end{figure}

In this scenario, we consider missing links between process entities that emerge when, e.g. new product variants are introduced that require a different production process. Furthermore, in a separate study we consider missing links from processes to their involved equipment entities, as shown in Figure \ref{fig:process}.

\subsection{Scenario: Observing New Events}
\label{ssec:events}

This scenario emerges frequently in automated factories in case the control logic that generates events of machines is modified, for example a new alarm message needs to be shown to the operator as soon as a certain oil pressure threshold is reached.

When new event entities are observed in the log that are missing annotations in the KG, the completion task can also be seen as a KG population task, since usually no previous information about a new event entity is present in the KG. Thus, predicting missing links means essentially introducing a completely new entity to the KG. 
This task corresponds to a well-known \textit{zero-shot} learning.

\begin{figure}[t!]
	\includegraphics[trim=0cm 0cm 0cm 0cm, clip, width=0.49\textwidth]{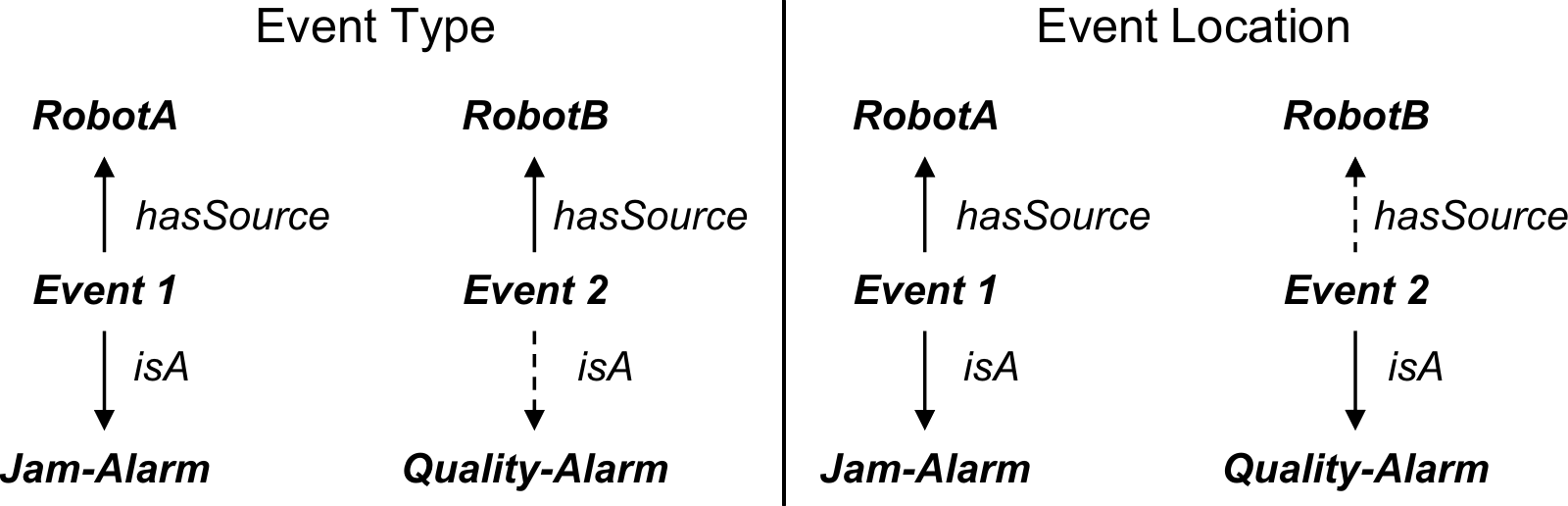}
	\caption{Event scenario, left: missing type information of event entities. Right: Missing source information of events}
	\label{fig:events}
\end{figure}

%


\section{Machine Learning Approach}
\label{sec:approach}

We now briefly describe our machine learning approach.

A \emph{knowledge graph} $\kg$ is a set of (positive) facts about a certain domain of interest represented as a set of triples of the form $\tuple{\mi{h,\;r,\;t}}$, where $\mi{h,t}\in \cE$ and $\mi{r}\in\cR$.
In our scenario the event-centric data is represented using a subset $\cX \subseteq \cE$ of entities from a given KG. 
A \emph{sequence} is an ordered set of (event) entities $s_i=(e_1,\dotsc,e_{m_i})$,
where $e_k\in\cX$.
A \emph{sequence dataset} $\seq$ is a set of sequences $\seq=\{s_1,\,\dotsc,\,s_n\}$.

Given a triple $\tuple{h, r, \_}$ (resp.~$\tuple{\_, r, t}$) with a missing entity, a KG $\kg$ and a sequence dataset $\seq$, the \emph{event-enhanced KG completion} is to utilize the background knowledge in $\seq$ to predict the missing $t$ (resp.~$h$) by retrieving a ranked list of possible candidate entities from a subset of all entities $\cE$.

Following the common practice, we solve the KG completion problem by
reducing it to a representation learning task,
whose main goal is to represent entities $h$, $t$ and 
relations $r$ occurring in $\kg$ in a low dimensional, e.g., $d$-dimensional,
vector space as vectors $\mathbf{h}, \mathbf{t}, \mathbf{r} \in \mathbb{R}^d$, which are referred to as \textit{embeddings}. 
In contrast to previously proposed extensions of the standard embedding approach, which improve the accuracy of the learned representations by taking into account additional knowledge in the textual form~\cite{Wang2016}, we make use of the background knowledge represented as sequences of events. Unlike text, sequence datasets reveal the exact order of event occurrences, they do not follow any grammatical rules, e.g., miss stop words and reflect the structure of the process that produced these sequences.

In this work we are looking for \emph{event-enhanced knowledge graph embeddings} to construct representations of $\mathbf{h}$, $\mathbf{t}$, $\mathbf{r}$ that leverage the sequential relationship between entities in $\seq$.
We note that despite the fact that $\seq$ is only directly connected to entities in $\cX$, a collective learning effect is introduced by incorporating event sequence information into the learning of KG embeddings that propagates event entity representations to other parts of the KG.

Our joint model combines the objective of KG embeddings $\cL_{\kg}$ 
and the objective of event sequence data embeddings $\cL_{\seq}$ 
using the joint formulation proposed in \cite{Xiao2017}:
\begin{align}
\label{eq:joint}
	\mathcal{L}_{joint} = \mathcal{L}_{\kg} + \alpha \mathcal{L}_{\seq},
\end{align}
where, $\alpha$ is a hyper-parameter used as a weighting factor.
Simultaneous training of both objective functions within an aggregated objective 
allows both models to influence each other through various parameter interconnections.

We consider three versions of $\mathcal{L}_{joint}$.

\begin{itemize}
	\item \emph{\EKLSkip} follows the intuition of the skipgram model \cite{Mikolov2013}, which relies  on the distributed representation hypothesis that a word is defined by its surrounding context. The goal of this model is to predict a context event given a particular center event in the log. 

	\item \emph{\EKLConcat} accounts for the sequential dependencies among events: it deals with the characteristics of short event sequences and preserves the information about their order when encoding entity embeddings; we achieve this by adapting a vector concatenation-based model that is similar to the paragraph-vector model \cite{Le.2014} without the notion of a paragraph. 

	\item \emph{\EKLRNN}
	employs a many-to-one vanilla RNN and feeds the $m-1$ predecessor 
	 events as inputs to make a prediction for the $m$-th event based on the last output state of the RNN.
\end{itemize}


\section{Evaluation}
\label{sec:system-evalusation}

\subsection{Evaluation Protocol}
We apply and evaluate the three novel approaches for event-enhanced KG completion, i.e., \EKLSkip, \EKLConcat and \EKLRNN, on each of the scenarios of Section \ref{sec:usecase}.
In each of the experiments, the original KG is split into a training, validation (10 \% of overall KG) and a test set that contains the artificially removed triples. Final model performance results are calculated based on the test set, for which we report a commonly-used evaluation metric:
\begin{itemize}
	\item \textbf{Mean Rank:} the average rank of the entities (head and tail) that would have been the correct ones.
\end{itemize}
The mean rank in our experiments corresponds to the \emph{filtered} version that has been originally proposed in \cite{Bordes2013}, i.e.  in the test set when ranking a particular triple $\tuple{h,r,t}$, all $\tuple{h,r,t'}$ triples with $t\neq t'$ are removed.
Employing grid search through the hyper-parameters we determine the best configuration by mean rank on the validation set with early stopping over a maximum of 100 epochs.

As a 
baseline, we use the default TransE model. 
For the incorporation of event sequence data, the skipgram model is also already a strong baseline in terms of comparison to order-preserving embeddings models.

\subsection{Results}
\begin{figure}[t!]
	\hspace{-5ex}
	\includegraphics[width=0.52\textwidth]{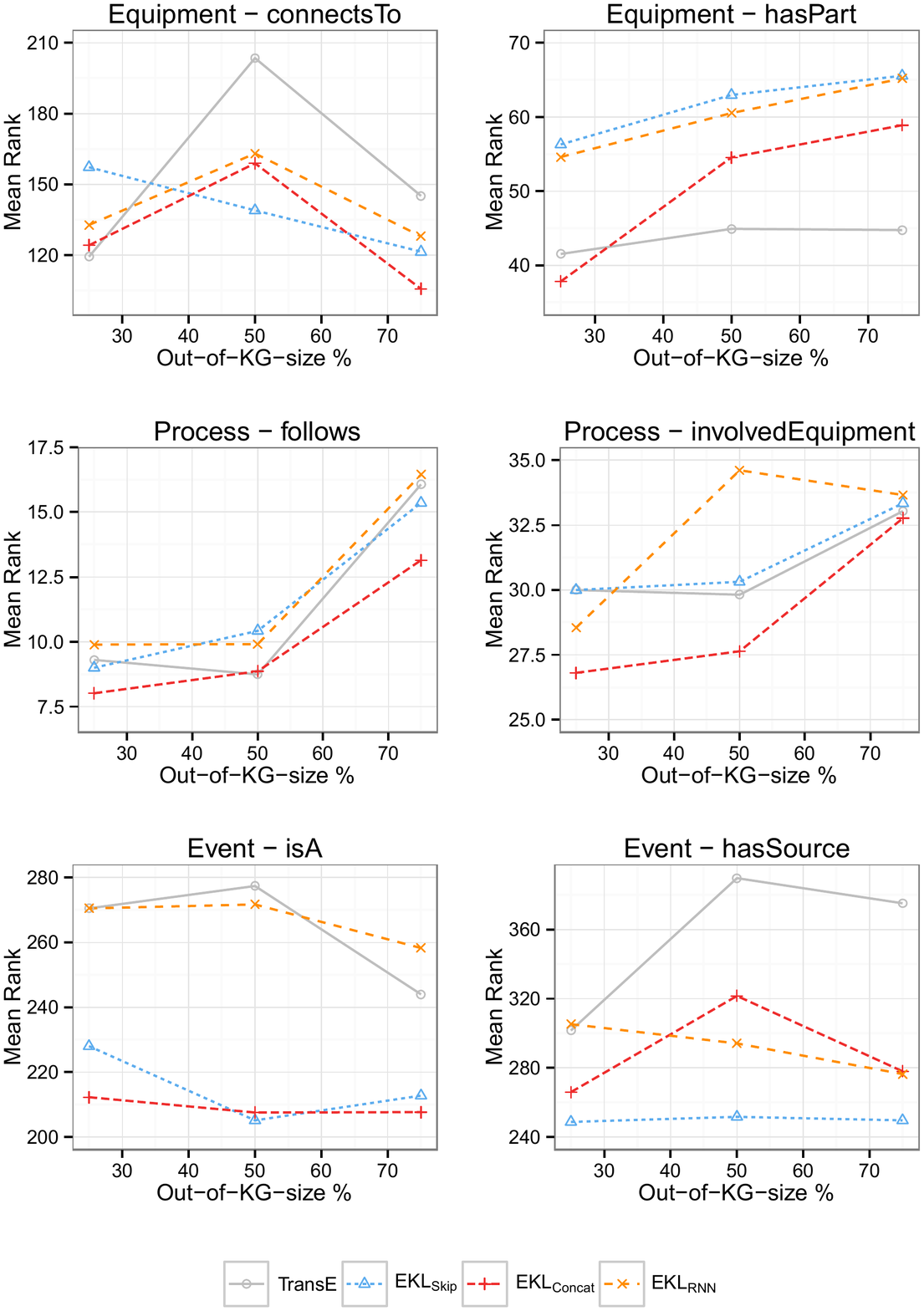}
	\caption{Mean rank statistics for KG completion task in each of the use case scenarios with increasing test set size}
	\label{fig:eval}
\end{figure}

In order to evaluate performance of our approach on the scenarios of Section \ref{ssec:equip}, 
we have designed dedicated test sets corresponding 
to the triples of the scenarios. 
For example, if triples with  \textit{connectedTo} relation are missing, then the test set contains a controlled proportion of all $\tuple{h, connectedTo, t}$ triples in the KG. 
We call this proportion the \textit{Out-of-KG-size} and varied it between 25\%, 50\% and 75\% for each relevant predicate. This way we can simulate the degree of missing links and therefore can assess how well the models can handle different amounts of missing information. 
The performance results with respect to mean rank are shown in Figure \ref{fig:eval}.
We now discuss them in details.

For the equipment connections scenario (top of Figure \ref{fig:eval}), it can be seen that incorporating events in the KG completion task does result in a lower mean rank compared to the standard TransE model as the proportion of missing connections is increased. On the other hand, this is not the case for the partonomy relation \textit{hasPart}, since the event-enhanced models' prediction quality is decreasing with growing Out-of-KG-size. 

For the new process scenario (middle of Figure \ref{fig:eval}), and  \textit{follows} and \textit{involvedEquipment} relations, especially \EKLConcat could robustly predict missing links with low mean rank. 
At the same time we observe that \EKLRNN and \EKLSkip in some parts perform worse than standard TransE. This supports our intuition for \EKLConcat that it can better capture process related relations.

For the new event scenario (bottom of Figure \ref{fig:eval}), one can see that, as expected, the event \textit{isA} and \textit{hasSource} relations benefit the most from incorporating their sequence as additional information. Most noticeably, 
\EKLSkip is robust to increasing size of missing links in both 
settings.


\section{Conclusion and Future Work}
\label{sec:conclusion}
In this paper we presented a concrete industrial scenario of a Siemens digital twin based on a knowledge graph that models a physical factory, including its equipments, processes, and also operational data, such as events. 
We showed how missing knowledge in this graph can be predicted by relying on machine learning that combines KGs with background data in the form of log files of events. 
In our scenarios the missing data corresponds to common changes in factories.
We evaluated our machine learning method in these scenarios and showed that with the help of our event-enhanced learning model it can do a good quality KG completion and therefore synchronise the digital and the physical representations of a smart factory. The KG completion performance results in most of the scenarios are promising and our models outperform a state-of-the-art KG completion model. Our approach performs best for population of KGs with new event entities.

\subsection*{Acknowledgements}
This work was partially supported by the EPSRC projects DBOnto,
MaSI$^3$, ED$^3$, and VADA.

\bibliographystyle{alpha}
\bibliography{library}

\newcommand{\etalchar}[1]{$^{#1}$}
\begin{thebibliography}{KMM{\etalchar{+}}17}

\bibitem[BUG{\etalchar{+}}13]{Bordes2013}
Antoine Bordes, Nicolas Usunier, Alberto Garc{\'{\i}}a{-}Dur{\'{a}}n, Jason
  Weston, and Oksana Yakhnenko.
\newblock Translating embeddings for modeling multi-relational data.
\newblock In {\em NIPS}, pages 2787--2795, 2013.

\bibitem[Dat16]{Datta2016}
Shoumen Datta.
\newblock {Emergence of Digital Twins}.
\newblock {\em arXiv:1610.06467}, 2016.

\bibitem[DHI12]{DBLP:books/daglib/0029346}
AnHai Doan, Alon~Y. Halevy, and Zachary~G. Ives.
\newblock {\em Principles of Data Integration}.
\newblock Morgan Kaufmann, 2012.

\bibitem[HBC17]{Hirmer2017}
Pascal Hirmer, Uwe Breitenb, and Ana Cristina.
\newblock {Automating the Provisioning and Configuration of Devices in the
  Internet of Things}.
\newblock {\em CSIMQ}, 9:28--43, 2017.

\bibitem[HGKW16]{DBLP:journals/internet/HorrocksGKW16}
Ian Horrocks, Martin Giese, Evgeny Kharlamov, and Arild Waaler.
\newblock Using semantic technology to tame the data variety challenge.
\newblock {\em IEEE Internet Computing}, 20(6):62--66, Nov 2016.

\bibitem[KGJ{\etalchar{+}}16]{Kharlamov2016a}
Evgeny Kharlamov, Bernardo~Cuenca Grau, Ernesto Jim{\'{e}}nez{-}Ruiz, Steffen
  Lamparter, Gulnar Mehdi, Martin Ringsquandl, Yavor Nenov, Stephan Grimm,
  Mikhail Roshchin, and Ian Horrocks.
\newblock Capturing industrial information models with ontologies and
  constraints.
\newblock In {\em {ISWC}}, pages 325--343, 2016.

\bibitem[KHS{\etalchar{+}}17]{OptiqueStatoil}
Evgeny Kharlamov, Dag Hovland, Martin~G. Skjæveland, Dimitris Bilidas, Ernesto
  Jiménez-Ruiz, Guohui Xiao, Ahmet Soylu, Davide Lanti, Martin Rezk, Dmitriy
  Zheleznyakov, Martin Giese, Hallstein Lie, Yannis Ioannidis, Yannis Kotidis,
  Manolis Koubarakis, and Arild Waaler.
\newblock Ontology based data access in {Statoil}.
\newblock {\em Journal of Web Semantics}, 44:3 -- 36, 2017.

\bibitem[KMM{\etalchar{+}}17]{OptiqueSiemens}
Evgeny Kharlamov, Theofilos Mailis, Gulnar Mehdi, Christian Neuenstadt, Özgür
  Özçep, Mikhail Roshchin, Nina Solomakhina, Ahmet Soylu, Christoforos
  Svingos, Sebastian Brandt, Martin Giese, Yannis Ioannidis, Steffen Lamparter,
  Ralf Möller, Yannis Kotidis, and Arild Waaler.
\newblock Semantic access to streaming and static data at {Siemens}.
\newblock {\em Journal of Web Semantics}, 44:54 -- 74, 2017.

\bibitem[KS{\"{O}}{\etalchar{+}}14]{Kharlamov2014}
Evgeny Kharlamov, Nina Solomakhina, {\"{O}}zg{\"{u}}r~L{\"{u}}tf{\"{u}}
  {\"{O}}z{\c{c}}ep, Dmitriy Zheleznyakov, Thomas Hubauer, Steffen Lamparter,
  Mikhail Roshchin, Ahmet Soylu, and Stuart Watson.
\newblock {How semantic technologies can enhance data access at siemens
  energy}.
\newblock In {\em ISWC}, pages 601--619, 2014.

\bibitem[LM14]{Le.2014}
Quoc Le and Tomas Mikolov.
\newblock {Distributed Representations of Sentences and Documents}.
\newblock In {\em ICML}, volume~32, pages 1188--1196, 2014.

\bibitem[MCCD13]{Mikolov2013}
Tomas Mikolov, Kai Chen, Greg Corrado, and Jeffrey Dean.
\newblock {Distributed Representations of Words and Phrases and their
  Compositionality}.
\newblock In {\em NIPS}, pages 1--9, 2013.

\bibitem[NMTG16]{Nickel2015}
Maximilian Nickel, Kevin Murphy, Volker Tresp, and Evgeniy Gabrilovich.
\newblock A review of relational machine learning for knowledge graphs.
\newblock {\em Proceedings of the {IEEE}}, 104(1):11--33, 2016.

\bibitem[RLBL15]{Ringsquandl2015}
Martin Ringsquandl, Steffen Lamparter, Sebastian-Philipp Brandt, and Raffaello
  Lepratti.
\newblock {Semantic-guided Feature Selection for Industrial Automation
  Systems}.
\newblock In {\em ISWC}. Springer, 2015.

\bibitem[RLLK17]{Ringsquandl.2017}
Martin Ringsquandl, Steffen Lamparter, Raffaello Lepratti, and Peer
  Kr{\"{o}}ger.
\newblock {Knowledge Fusion of Manufacturing Operations Data using
  Representation Learning}.
\newblock In {\em IFIP APMS}. Springer, 2017.

\bibitem[RVLB15]{Rosen2015}
Roland Rosen, Georg {Von Wichert}, George Lo, and Kurt~D. Bettenhausen.
\newblock {About the importance of autonomy and digital twins for the future of
  manufacturing}.
\newblock {\em IFAC-PapersOnLine}, 28(3):567--572, 2015.

\bibitem[WL16]{Wang2016}
Zhigang Wang and Juan-Zi~Juanzi Li.
\newblock {Text-Enhanced Representation Learning for Knowledge Graph}.
\newblock {\em IJCAI}, pages 1293--1299, 2016.

\bibitem[XHMZ17]{Xiao2017}
Han Xiao, Minlie Huang, Lian Meng, and Xiaoyan Zhu.
\newblock {SSP: Semantic Space Projection for Knowledge Graph Embedding with
  Text Descriptions}.
\newblock {\em AAAI}, pages 1--10, 2017.

\end{thebibliography}

\end{document}